\documentclass{article}

\usepackage{arxiv}

\usepackage[utf8]{inputenc} 
\usepackage[T1]{fontenc}    
\usepackage{hyperref}       
\usepackage{url}            
\usepackage{booktabs}       
\usepackage{amsfonts}       
\usepackage{nicefrac}       
\usepackage{lipsum}		
\usepackage{graphicx}
\usepackage{natbib}
\usepackage{doi}

\title{Machine Psychology: Integrating Operant Conditioning with the Non-Axiomatic Reasoning System for Advancing Artificial General Intelligence Research}

\date{} 					

\author{ \href{https://orcid.org/0000-0001-5547-3866}{\includegraphics[scale=0.06]{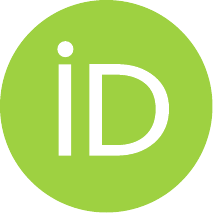}\hspace{1mm}Robert ~Johansson} \\
	Department of Psychology\\
	Stockholm University\\
	Stockholm, Sweden \\
	\texttt{robert.johansson@psychology.su.se} \\
}

\renewcommand{\headeright}{}
\renewcommand{\undertitle}{}
\renewcommand{\shorttitle}{Machine Psychology}

\hypersetup{
pdftitle={Machine Psychology},
pdfsubject={q-bio.NC, q-bio.QM},
pdfauthor={Robert ~Johansson},
pdfkeywords={First keyword, Second keyword, More},
}

\begin{document}
\maketitle

\begin{abstract}
This paper presents an interdisciplinary framework, Machine Psychology, which integrates principles from operant learning psychology with a particular Artifical Intelligence model, the Non-Axiomatic Reasoning System (NARS), to advance Artificial General Intelligence (AGI) research. Central to this framework is the assumption that adaptation is fundamental to both biological and artificial intelligence, and can be understood using operant conditioning principles. The study evaluates this approach through three operant learning tasks using OpenNARS for Applications (ONA): simple discrimination, changing contingencies, and conditional discrimination tasks. 

In the simple discrimination task, NARS demonstrated rapid learning, achieving 100\% correct responses during training and testing phases. The changing contingencies task illustrated NARS's adaptability, as it successfully adjusted its behavior when task conditions were reversed. In the conditional discrimination task, NARS managed complex learning scenarios, achieving high accuracy by forming and utilizing complex hypotheses based on conditional cues. 

These results validate the use of operant conditioning as a framework for developing adaptive AGI systems. NARS’s ability to function under conditions of insufficient knowledge and resources, combined with its sensorimotor reasoning capabilities, positions it as a robust model for AGI. The Machine Psychology framework, by implementing aspects of natural intelligence such as continuous learning and goal-driven behavior, provides a scalable and flexible approach for real-world applications. Future research should explore using enhanced NARS systems, more advanced tasks and applying this framework to diverse, complex tasks to further advance the development of human-level AI.

\end{abstract}


\section{Introduction}

Artificial General Intelligence (AGI) is the task of building computer systems that are able to understand or learn any intellectual task that a human being can. This type of AI is often contrasted with narrow or weak AI, which is designed to perform a narrow task (e. g., facial recognition or playing chess). There are several diverse research approaches to AGI including brain-based approaches \citep[e.g., ][]{hawkins2021thousand}, projects that aims to implement different cognitive functions separately \citep[e.g., ][]{laird2019soar}, and principle-based approaches \citep[e.g., ][]{wang2013nalbook, hutter2004universal}. Recently, Large Language Models like GPT-4 have also been introduced as a potential pathway towards achieving more generalizable AI systems \citep{bubeck2023gpt4}.

One major challenge in contemporary AGI research is the lack of coherent theoretical frameworks \citep{wang2012theories, wang2019defining}. This scarcity of unified models to interpret and guide the development of AGI systems seems to have led to a fragmented landscape where researchers often work in isolation on narrowly defined problems. Coherent research frameworks could also provide a roadmap and evaluation criteria for AGI development, fostering more collaborative and interdisciplinary efforts. The fact that AGI research has not progressed as rapidly as some had hoped could very well be attributed to the lack of these comprehensive frameworks and the absence of standardized benchmarks for measuring progression towards AGI capabilities.

This work aims to address this challenge by proposing a novel framework that outlines key milestones and metrics for evaluating progress in the field of artificial general intelligence (AGI). A fundamental assumption in this work is that \textit{adaptation} is at the heart of general intelligence. Adaptation is typically divided in \textit{ontogenetic adaptation}, which involves the changes that occur with an organism over its lifespan, and 2) \textit{phylogenetic adaptation}, which refers to the evolutionary changes that occur across generations within a species.

\textit{Learning} has within the field of learning psychology, been equated with ontogenetic adaptation, where an individual's experiences directly impact its capabilities and behaviors \citep{dehouwer2013learning}. \textit{Operant conditioning} is one type of learning that involves adaptation in the form of behavioral changes due to consequences of actions. Given the enormous amount of empirical progress generated by operant conditioning research in learning psychology, the principle of operant conditioning and its associated research tradition could be a guiding principle for AGI research.

One particular approach to building AGI is the Non-Axiomatic Reasoning System (NARS) \citep{wang2013nalbook, wang2022intelligence}. NARS is an adaptive reasoning system that operates on the principle of insufficient knowledge and resources, a condition that is often true for real-world scenarios. Hence, NARS is a principle-based approach that aims to address the challenges of building AGI systems that can operate effectively in dynamic and unpredictable environments \citep{wang2019defining}. There are several NARS implementations available. One implementation is OpenNARS for Applications (ONA), that is designed to provide a practical framework for integrating NARS into various applications, with a particular focus on robotics \citep{hammer2020opennars, hammer2023transbot}. ONA is built with sensorimotor reasoning at its core, enabling it to process sensory data in real-time and respond with appropriate motor actions. Sensorimotor reasoning, as implemented in ONA, permits the system to make sense of the world much in the same way as biological organisms do, by directly interacting with its environment and learning from these interactions. The fact that NARS systems are focused on adaptation, and that ONA has a strong emphasis on sensorimotor capabilities, suggests that they are particularly well-suited for implementing the principle of operant conditioning.

This work presents Machine Psychology, an interdisciplinary framework for advancing AGI research. It integrates principles from learning psychology, with the theory and implementation of NARS. Machine Psychology starts with the assumption that adaptation is fundamental to intelligence, both biological and artificial. As it is presented here, Machine Psychology is guided by the theoretical framework of learning psychology, and the principle of operant conditioning in particular. A sensorimotor-only version of ONA \citep{hammer2022reasoning} is used to demonstrate the feasibility of using these principles to guide the development of intelligent systems. One way to describe the integration of operant conditioning and NARS presented in this paper is that the ability to learn and adapt based on feedback from the environment, is implemented using sensorimotor reasoning that is the core of ONA. An analogy is that a neurobiological explanation of operant conditioning could be argued to be part of a biological basis for adaptive behaviors observed in many species \citep{brembs2003operant}, similarly, the implementation within ONA using temporal and procedural inference rules offers an alternative explanation of the core of adaptive behavior and cognition.

We evaluate the Machine Psychology framework by carrying out three operant learning tasks with NARS. The first is a simple discrimination task in where NARS needs to learn, based on feedback, to choose one stimulus over another, demonstrating a fundamental aspect of learning based on the consequences of actions. The second task is more complex than the first in that the conditions of the experiment is changed midway through the task, requiring NARS to adapt its choice strategy based on the new conditions. The third experiment is a conditional discrimination task, where NARS is presented with pairs of stimuli, and must learn to select the correct stimulus based on a conditional cue that changes throughout the task, requiring an increased level of adaptability. Methods from learning psychology are used to design the experiments and guide the evaluation of the results. We explain the results by describing how the sensorimotor reasoning used by ONA enables it to adaptively modify its behavior based on the consequences of its actions. The Machine Psychology framework is demonstrated to provide a coherent experimental approach to studying the core of learning and cognition with artificial agents, and also offers a scalable and flexible framework that potentially could significantly advance research in Artificial General Intelligence (AGI).

The paper is organized as follows. Section 2 presents a background on the principles of operant conditioning and its significance in learning psychology. Section 3 introduces NARS with a focus on its foundational concepts. Section 4 describes the architecture of OpenNARS for Applications with a particular focus on its sensorimotor reasoning abilities. Section 5 discusses related work to our approach. Section 6 presents the Machine Psychology framework and how it integrates operant conditioning principles with NARS. Section 7 describes the details of the methods and experimental setup used in the evaluation of our approach. Section 8 presents the results from the experiments. Section 9 concludes the paper and outlines how the Machine Psychology framework could be used to further advance the field of AGI.

\section{Operant conditioning} \label{section:operant}

The work presented in this paper takes a \textit{functional} approach to learning and adaptation \citep{dehouwer2020book}, and to science in general. Such approach to learning is rooted in the principles of behaviorism, which emphasizes the role of environmental interactions in shaping behavior, rather than mechanistic explanations of how internal processes affect behavior. It stems particularly from the work of B. F. Skinner, who laid much of the groundwork for understanding how consequences of an action affect the likelihood of that action being repeated in the future \citep{skinner1938behavior}. Skinner was influenced by physicist and philosopher Ernst Mach, who emphasized the use of \textit{functional relations} in science to describe relations between events, rather than using a traditional mechanistic causal framework \citep{chiesa1994radical}.

In the functional learning psychology tradition, learning (as ontogenetic adaptation) is defined as \textit{a change in behavior due to regularities in the environment} \citep{dehouwer2013learning}. Several types of learning can be classified under this perspective. Operant conditioning is defined as \textit{a change in behavior due to regularities between behavior and stimuli} \citep{dehouwer2013learning}. Other types of learning can be defined based on other the regularities in operation. A few comments regarding these definitions of learning and operant conditioning follow, as clarified by \cite{dehouwer2013learning}. First, in line with \cite{skinner1938behavior} \textit{behavior} is defined very broadly, encompassing any observable action or response from an organism. This includes responses that are only in principle observable, such as internal physiological changes, neural processes or cognitive events. In addition to this, behavior is defined to always be a function of one or more stimuli, while a response is just an observable reaction. Second, \textit{regularities} are defined to be any patterns of events or behavior that go beyond a single occurrence. This can be the same events happening repeatedly, or two or more events or behaviors happening at the same time. Third, the definition signals a particular view of causality (“due to”). As highlighted above, this research tradition emphasizes \textit{functional relations} between environmental regularities and changes in behavior. This implies that learning, from this perspective, cannot be directly observed, but must be inferred from the systematic changes in behavior in response to modifications in the environment \citep{dehouwer2013learning}. Such inferences do depend on an observer, whose scientific goals and theoretical orientations shape the interpretation. An example follows that aims to clarify this definition further.

\subsection{An example of operant conditioning}

Imagine a rat in an experimental chamber used to study behavior. The chamber contains a small loudspeaker, a lever that can be pressed, and a water dispenser, where the delivery of water is controlled by the researcher. The researcher aims to shape the rat into turning around when techno music is played, and press the lever when classical music is played. To do this, the researcher uses an operant conditioning procedure. Before the experiment, the rat has been deprived of water for a short period. Initially, any tendency to turn around when techno music is played might be followed by the delivery of water. This makes the behavior more likely to occur in the future under similar conditions. Conversely, when classical music is played, and the rat presses the lever, the delivery of water follows behavior as well. Over time, the rat learns to turn around or press the lever based on the type of music that is playing. This illustrates operant conditioning, more specifically \textit{positive reinforcement} (to be defined below). A video of a rat performing in this experiment can be found online \citep{wmu2}.

This example can be considered an effective demonstration of operant conditioning. The behavior in this example is a function of both the music and the water. It illustrates the point made above that behavior studied from this perspective is not just an isolated motor action, but is also significantly influenced by the surrounding environment and the consequences that follow the behavior. The regularities in operation are reoccurring patterns of behavior and stimuli, for example lever pressing and the delivery of water, but also a regularity regarding classical music and lever pressing. To make a causal statement about learning, we would need to observe the rat before and after interacting with these stimuli. Before the interaction, we might for example observe the rat exploring the cage or do random actions when different types of music was playing. After the interaction however, if we observe the behaviors described in the example, this would be a clear change in behavior - from for example cage exploring when classical music is playing to lever pressing when the same music is playing. If we could argue that the change in behavior is due to the procedural arrangements (regularities), then we could potentially claim that this qualifies as an instance of learning. The type of learning it would indicate is operant conditioning since the regularities involved were between responses and stimuli (rather than for example a repeated pairing of stimuli as with classical conditioning). More specifically, it would be an instance of \textit{reinforcement}, a kind of operant conditioning that involves an increase in target behavior due to the consequences.

\subsection{Three levels of analysis}

A learning situation such as the one illustrated in the example can be analyzed on three levels using learning psychology: 1) The descriptive level (or level of procedure), 2) The functional level (or the level of effect), and 3) The cognitive level (or the level of mechanism) \citep{dehouwer2020book}. At the descriptive level, the procedural arrangement is from the perspective of the researcher. The different sounds is used to signal if a relationship between behaving in a certain way and water holds. This is a description of procedures initiated by the researcher. It doesn’t mean that the rat has learned based on these arrangements. The functional level however, is closer to describing the relations from the rat’s perspective. If the rat turns around if and only if the techno music is playing, then that music functions as a cue for that behavior. Similarly, it is only if the delivery of water has an effect on the behavior, that it functions as a reinforcer. If there is no change in lever pressing or turning due to the water being delivered, then that consequence has no effect. Finally, the cognitive level can be used to describe certain mental mechanisms, like association formations, that could explain how the operant learning processes take place \citep{dehouwer2020book}. 

The importance of distinguishing these levels can’t be overstated. Just using an operant conditioning procedure (like the one above), does not mean that the subject learns in the form of operant conditioning. When doing functional learning research, we arrange procedures and study the effect on behavior change. Learning, from this perspective, is hence defined on the functional level. As stated above, a term such as reinforcement is also defined as an effect rather than a mechanism. This also means that explanations on the cognitive/mechanistic level are not part of the learning definition \citep{dehouwer2013learning}. It also opens up for different kinds of explanations in terms of mechanisms - for example propositional networks as something different from association-based learning theories.

\subsection{The three-term contingency} \label{section:threeterm}

To describe behavior as an interaction between the organism and its environment, Skinner introduced the concept of the three-term contingency, which consists of discriminative stimulus (Sd), response (R), and resultant stimulus (Sr) \citep{skinner1953science, dehouwer2020book}. Sometimes the terms antecedent, behavior, consequence is used to reflect the same triadic relationship in a more general context. The example above describes a contingency of reinforcement, in where the music functioned as a discriminative stimulus, and the water functioned as a resultant stimulus of the reinforcing type. Arguably, a fourth term is needed to account for the fact that it is only when the rat is water-deprived that the water serves as a reinforcing stimulus. Exposing the rat to water deprivation is therefore a critical antecedent condition, which modifies the efficacy of the water as a reinforcer. Such antecedent procedural arrangement is typically called an \textit{establishing operation} (EO) (which is an example of a \textit{motivating operation}. Hence, the EO could be included as a fourth term in the three-term contingency description. 

In the example above, we could also have imagined enhancing the procedure with adding a light that could be on or off, signaling if the relation between music type, the rat's behavior, and water would hold or not. Since the function of the other terms would be conditional on the state of the light, a stimulus that functions in this way is called a \textit{conditional discriminative stimulus} \citep{lashley1938conditional}. 

\subsection{Where is the organism?}

In a functional analysis of behavior, it is an interaction between organism and environment that is being analyzed. Hence, it is not the organism itself in isolation that is of interest. Technically, it is interactions between stimulus functions and response functions that are being studied, for example an interaction between seeing a lever and pressing it, or hearing techno music and reacting to it. There are other conceptual schemas in functional learning psychology than the three-term contingency that takes into account the complexity of these interactions \citep{hayesfryling2018}, but for this paper, what has been presented above is a sufficient conceptual framework. Importantly though, in functional learning research, these relations between procedural arrangements and behavior change does depend on an organism in that they enable response functions. This does not in an way mean that the organism causes behavior. Rather, the organism can be considered a \textit{participant} in the arrangements \citep{rochebarnes1997behavior}. 

Recently, \cite{dehouwer2022learning} extended their conceptual work on learning beyond that of organisms, for example to also incorporate the study of learning with genes, groups, and machines. At the descriptive level of analysis, they replace the term response with that of \textit{state transition}. In the example above with the rat, a change from exploring the cage to pressing the lever could be described as such a state transition (moving from a state of exploration to a state of lever pressing). Importantly, states, as defined from this perspective, are used to describe state transitions. A behavior is then defined as a state transition in relation to one or more stimuli, for example in relation to the music being played. This is once again a functional definition - the behavior is a function of stimuli. Learning is still defined as changes in behavior (state transitions in relation to stimuli) that occurs due to regularities \citep{dehouwer2022learning}. 

Based on this conceptual change a \textit{system} is defined as a construct from the perspective of an observer, as sets of states that can be used to describe change. In common language we often refer to a rat “being a system”, but technically it is rather that the system is a collection of topographical descriptions of a rat’s physiological responses. With computer systems they could typically be described as collections of interdependent systems \citep{hayesfryling2018}. For example a robot taking part of an experimental task, might for example be described as hardware movements (like the robot arm) that are dependent on sensory equipment and on software interpreting those sensory inputs. 

\subsection{Human-level intelligence from an operant perspective}

All above examples involve learning that are animal-level in the sense that they could be observed with an animal like a rat. The fact that operant conditioning can be observed with both humans and animals does not however mean that these processes are irrelevant for achieving human-level intelligence with artificial systems. On the contrary, we would argue that mechanisms enabling operant conditioning at the core \citep[as with OpenNARS for Applications;][]{hammer2022reasoning} could very well be integral to the development of complex cognitive behaviors with AGI systems. While outside the scope of the present paper, a brief account of human-level intelligence from an operant perspective follows. 

In functional learning psychology, there are theories regarding what kinds of extensive learning histories an organism needs to develop sophisticated cognitive capabilities. The functional learning theory Relational Frame Theory \citep[RFT;][]{rft2001} argue for the necessity of histories involving patterns of relating leading to derived performances that are not directly taught but are instead inferred from these relational patterns. Importantly, learning to derive relations in accordance with for example similarity, opposition, comparision, etc, enables the development of complex cognitive skills such as language understanding, problem-solving, and abstract reasoning. Such patterns of relating (called “relational frames” in RFT) are assumed to be operant behaviors in themselves, which are learned through interaction with the environment and are subject to reinforcement \citep{hayes2021relating}. Hence, from the perspective of RFT, intelligence as a whole can be viewed as a collection of learned patterns of relating, which are not static but dynamically evolve through continuous interaction with one's environment. This means, that a roadmap towards human-level AI from the perspective of functional learning psychology and RFT, would clearly emphasize operant conditioning abilities at the core \citep{johansson2019aarr, johansson2020scientific}.

\section{Non-Axiomatic Reasoning Systems} \label{section:nars}

A Non-Axiomatic Reasoning System (NARS) is a type of artificial intelligence system that operates under the assumption of insufficient knowledge and resources (AIKR) \citep{wang1995thesis, wang2006rigid, wang2013nalbook}. The AIKR principle dictates that the system must function effectively despite having limited information and computational resources, a scenario that closely mirrors real-world conditions and human cognitive constraints.

All NARS systems implement a Non-Axiomatic Logic (NAL) \citep{wang2013nalbook}, a term logic designed to handle uncertainty using experience-grounded truth values. Most NARS systems also makes use of concept-centric memory structure, which organizes the system's memory based on terms and subterms from the logic statements, leading to a more effective control of the inference process. Furthermore, all NARS systems use a formal language \textit{Narsese}, that allows encoding of complex information and communication of NAL sentences within and between NARS systems.

\subsection{Core principles of NARS}

NARS systems are built on a few key concepts that distinguish them from traditional AI systems \citep{wang2022intelligence}:

\begin{enumerate}

\item \textbf{Adaptation Under AIKR: } Unlike systems that assume abundant knowledge and resources, NARS thrives under constraints. It manages finite processing power and storage, operates in real-time, and handles tasks with varying content and urgency. This adaptability ensures that NARS remains relevant in dynamic and unpredictable environments.

\item \textbf{Experience-Based Learning and Reasoning: } Central to NARS is its concept-centered representation of knowledge. Concepts in NARS are data structures with unique identifiers, linked through relations such as inheritance, similarity, implication, and equivalence. These relationships are context-sensitive and derived from the system's experiences, allowing NARS to continuously update and refine its knowledge base as it encounters new information.

\item \textbf{Non-Axiomatic Logic: } Traditional AI often relies on axiomatic systems where certain truths are taken as given. In contrast, NARS employs non-axiomatic logic, where all knowledge is subject to revision based on new experiences. This approach supports a variety of inference methods, including deduction, induction, abduction, and analogy, enabling NARS to reason in a manner that is both flexible and grounded in empirical evidence.

\end{enumerate}

\subsection{Problem-Solving and Learning} 

NARS processes three types of tasks: incorporating new knowledge, achieving goals, and answering questions. It uses both forward and backward reasoning to handle these tasks, dynamically allocating its limited resources based on task priorities. This approach, known as case-by-case problem-solving, means that NARS does not rely on predefined algorithms for specific problems. Instead, it adapts to the situation at hand, providing solutions that are contextually appropriate and continuously refined.

Learning in NARS is a self-organizing process \citep{wang2022intelligence}. The system builds and adjusts its memory structure—a network of interconnected concepts—based on its experiences. This structure evolves over time, allowing NARS to integrate new knowledge, resolve conflicts, and improve its problem-solving capabilities. Unlike many machine learning models that require large datasets and extensive training, NARS learns incrementally in interaction with its environment and can accept inputs at various levels of abstraction, from raw sensorimotor data to complex linguistic information.

\subsection{A Unified Cognitive Model}

One of the most significant aspects of NARS is its unified approach to cognitive functions. In NARS, reasoning, learning, planning, and perception are not separate processes but different manifestations of the same underlying mechanism. This integration provides a coherent framework for understanding and developing general intelligence, making NARS a versatile tool for a wide range of AI applications \citep{wang2022intelligence}.

In conclusion, NARS represents a significant departure from conventional AI paradigms by embracing the challenges of limited knowledge and resources. Its unique combination of non-axiomatic reasoning, experience-based learning, and adaptive problem-solving positions NARS as a robust model for advancing artificial general intelligence.

\section{OpenNARS for Applications} \label{section:ona}

OpenNARS for Applications (ONA) is a highly effective implementation of a NARS, designed to be suitable for practical applications such as robotics \citep{hammer2020opennars}. At the core of ONA lies sensorimotor reasoning, which integrates sensory processing with motor actions to enable goal-directed behavior under conditions of uncertainty and limited resources. ONA differs from other NARS systems in several key aspects, including: 

\begin{enumerate}
\item \textbf{Event-Driven Control Process:} ONA incorporates an event-driven control mechanism that departs from the more probabilistic and bag-based approach used in traditional NARS implementations, such as OpenNARS \citep{lofthouse2019alann}. This shift allows ONA to prioritize processing based on the immediacy and relevance of incoming data and tasks. The event-driven approach is particularly advantageous in dynamic environments where responses to changes must be timely and context-sensitive.

\item \textbf{Separation of Sensorimotor and Semantic Inference:} Unlike other NARS models that often blend various reasoning functions, ONA distinctly separates sensorimotor inference from semantic inference \citep{hammer2020opennars}. This division allows for specialized handling of different types of reasoning tasks—sensorimotor inference can manage real-time, action-oriented processes, while semantic inference deals with abstract, knowledge-based reasoning. This separation helps to optimize processing efficiency and reduces the computational complexity involved in handling diverse reasoning tasks simultaneously.

\item \textbf{Resource Management:} ONA places a strong emphasis on managing computational resources effectively, adhering to the Assumption of Insufficient Knowledge and Resources (AIKR). It is designed to operate within strict memory and processing constraints, employing mechanisms like priority-based forgetting and constant-time inference cycles. These features ensure that ONA can function continuously in resource-limited settings by efficiently managing its cognitive load and memory usage.

\item \textbf{Advanced Data Structures and Memory Management:} ONA utilizes a sophisticated system of data structures that include events, concepts, implications, and a priority queue system for managing these elements. This setup facilitates more refined control over memory and processing, prioritizing elements that are most relevant to the system's current goals and tasks. It also helps in maintaining the system’s performance by managing the complexity and volume of information it handles.

\item \textbf{Practical Application Focus:} The architectural and control changes in ONA are driven by a focus on practical application needs, which demand reliability and adaptability. ONA is tailored to function effectively in real-world settings that require autonomous decision-making and adaptation to changing environments, making it more applicable and robust than its predecessors for tasks in complex, dynamic scenarios \cite{hammer2020opennars}.
\end{enumerate}

\subsection{The architecture of ONA}

\begin{figure}[h!]
\begin{center}
\includegraphics[scale=0.6]{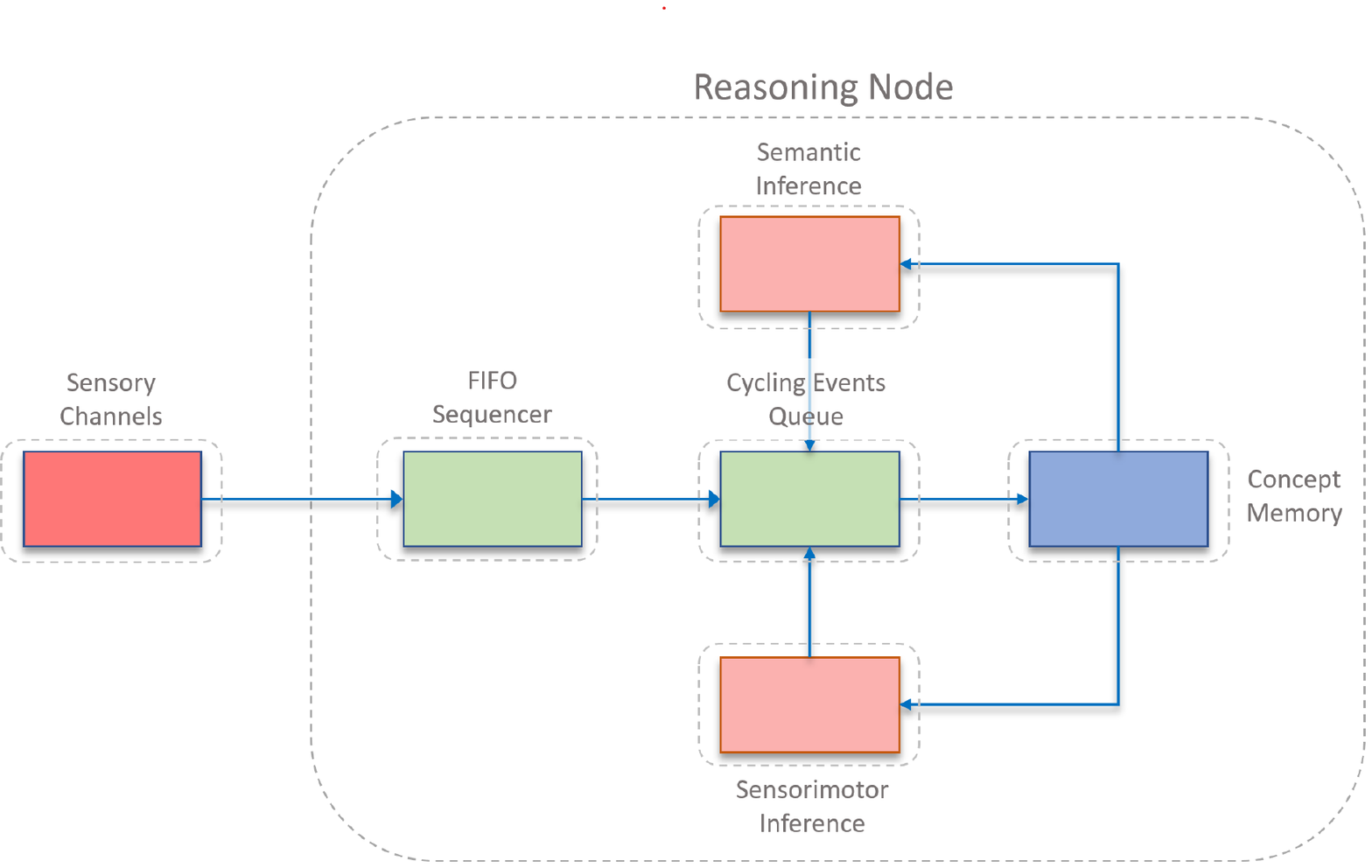}
\end{center}
\caption{
An overview of the architecture in OpenNARS for Applications (ONA). Reprinted with permission from Patrick Hammer, the author of ONA.
} 
\label{fig_ona}
\end{figure}

ONA’s architecture is composed of several interrelated components that work together to process sensory input, manage knowledge, make decisions, and learn from experience. These components are designed to handle the dynamic and uncertain nature of real-world environments, ensuring that the system can adapt and respond effectively \citep{hammer2022reasoning}. The architecture is illustrated in Figure \ref{fig_ona}. ONA's architecture has a number of key components: 1) Event Providers, 2) FIFO Sequencer, 3) Cycling Events Queue, 4) Concept Memory, 5) Sensorimotor Inference Block, and 6) Declarative Inference Block. Each of these components plays a crucial role in ONA’s operation, as described in detail below.

\subsubsection{Event providers}
Event providers are responsible for processing sensory inputs from various modalities, converting raw data into structured statements that the reasoning system can interpret. Each event provider is specialized for different types of sensory information, such as visual, auditory, or tactile data. These providers ensure that all relevant environmental information is captured and encoded as events, which are then fed into the system for further processing. The main functionality can be summarized as follows:

\begin{itemize}
	\item \textbf{Sensor Data Processing: } Event providers preprocess raw sensor data to filter noise and extract meaningful information.
	\item \textbf{Event Encoding: } The processed data is encoded into statements or events that can be understood by the ONA system.
\end{itemize}

\subsubsection{FIFO Sequencer}

The FIFO (First-In-First-Out) Sequencer maintains a sliding window of recent events. This component is essential for building and strengthening temporal implication links, which are used to understand the sequence of events and their relationships over time. By keeping track of the recent history, the FIFO Sequencer allows ONA to form hypotheses about temporal patterns and causal relationships. As a note, in recent versions of ONA, the FIFO was removed and replaced by an explicit temporal inference block. This design is however not yet described in any scientific publications, and therefore the design with the FIFO has been described. The main functionality can be summarized as follows:

\begin{itemize}
	\item \textbf{Event Sequencing: } It organizes events in a chronological order, maintaining a window of the most recent events.
	\item \textbf{Temporal Implications: } Builds and strengthens links between events based on their temporal proximity and sequence.
\end{itemize}

\subsubsection{Cycling Events Queue}

The Cycling Events Queue is a priority queue that serves as the central attention buffer of the system. All input and derived statements enter this queue, but only a subset can be selected for processing within a given timeframe due to the fixed capacity of the queue. This mechanism ensures that the most relevant and urgent information is processed first, while less critical information is discarded or delayed. The main functionality can be summarized as follows:

\begin{itemize}
	\item \textbf{Priority Management: } Events are prioritized based on their importance and relevance to current goals.
	\item \textbf{Attention Focus: } Ensures that the system’s limited processing resources are focused on the most critical tasks.
\end{itemize}

\subsubsection{Concept Memory}

Concept Memory acts as the long-term memory of the ONA system. It stores temporal hypotheses and supports their strengthening or weakening based on prediction success. This memory component allows ONA to retain knowledge over long periods, enabling cumulative learning and the ability to recall past experiences to inform current decision-making. The main functionality can be summarized as follows:

\begin{itemize}
	\item \textbf{Hypothesis Management: } Stores and manages temporal and procedural hypotheses about the environment.
	\item \textbf{Evidence Accumulation: } Strengthens or weakens stored hypotheses based on new evidence and prediction outcomes.
\end{itemize}

\subsubsection{Sensorimotor Inference Block}

The Sensorimotor Inference Block is responsible for handling decision-making and subgoaling processes for goal events selected from the Cycling Events Queue. This component invokes algorithms for goal achievement, generating actions or subgoals that guide the system’s behavior towards fulfilling its objectives. The main functionality can be summarized as follows:

\begin{itemize}
	\item \textbf{Decision Making: } Selects the best actions to achieve current goals based on stored knowledge and recent events.
	\item \textbf{Subgoaling: } Decomposes complex goals into manageable subgoals, facilitating step-by-step achievement of objectives.
\end{itemize}

\subsubsection{Declarative Inference Block}

The Declarative Inference Block is responsible for higher-level reasoning tasks such as feature association, prototype formation, and relational reasoning. It utilizes human-provided knowledge to enhance the system’s understanding of the environment and improve its reasoning capabilities. Though not utilized in the specific experiments described in the paper, this block is crucial for applications requiring complex knowledge integration and abstract reasoning. The main functionality can be summarized as follows:

\begin{itemize}
	\item \textbf{Feature Association: } Links related features and concepts based on observed patterns and external knowledge.
	\item \textbf{Relational Reasoning: } Understands and reasons about relationships between different concepts and entities.
\end{itemize}

\subsection{The operations of ONA} \label{section:ops_ona}

The main operations of ONA will be described below.

\subsubsection{Truth Value Calculation}

Truth values in ONA are based on positive and negative evidence supporting or refuting a statement, respectively. The system uses two measures: frequency (the ratio of positive evidence to total evidence) and confidence (the ratio of total evidence to total evidence plus one). This approach allows ONA to represent degrees of belief, accommodating the inherent uncertainty in real-world information.

The calculation of frequency and confidence is conducted as follows. Frequency: $f = \frac{w^+}{w}$, and Confidence: $c = \frac{w}{w + 1}$, where $w$ is the total amount of evidence, and $w^+$ is the positive evidence.

These values are used to evaluate the truth of implications and guide decision-making processes, ensuring that actions are based on the most reliable and relevant information available.

\subsubsection{Implications and Learning}

ONA forms temporal and procedural implications through induction and revises them based on new evidence. Temporal implications represent sequences of events, while procedural implications represent action-outcome relationships. Learning involves accumulating positive and negative evidence for these implications and adjusting their truth values accordingly. Also if an implication exists (for example \verb!<(<A1 --> [left]> &/ ^left) =/> G>!, and $A1$ was observed followed by \verb!^left!, an assumption of failure will be applied to the implication for implicit anticipation. This means, if the anticipation fails, the truth of the implication will be reduced by the addition of negative evidence, via an implicit negative $G$ event, while the truth will increase due to positive evidence in case $G$ happened.

The learning process at the core consists of:

\begin{itemize}
	\item \textbf{Event Sequences: } Implications are formed when related events occur within the sliding window maintained by the FIFO Sequencer.
	\item \textbf{Evidence Update: } Positive and negative evidence is accumulated and used to revise the truth values of implications, ensuring they reflect the system’s experiential knowledge.
\end{itemize}

\subsubsection{Decision Making and Subgoaling}

ONA’s decision-making process is goal-driven, leveraging its knowledge of temporal and procedural implications to select actions or generate subgoals. The system evaluates the desire value of goals and subgoals, prioritizing them based on their likelihood of success and relevance to current objectives. The decision process can be described as follows:

\begin{itemize}
	\item \textbf{Goal Deducation and Evaluation: } Determines the most desirable actions or subgoals based on stored implications and recent events.
	\item \textbf{Subgoal Generation: } Breaks down complex goals into smaller, manageable subgoals, facilitating efficient achievement through step-by-step actions. 
\end{itemize}

\subsubsection{Motor Babbling}

To trigger executions when no procedural knowledge yet exists, ONA periodically invokes random operations, a process called \textit{Motor Babbling}. This enables ONA to execute operations despite any procedural knowledge that applies. Without this ability, ONA would not be able to do its initial steps of learning procedural knowledge \citep{hammer2020opennars}.

\subsection{Conclusion}

The architecture of ONA integrates various components that collectively enable it to reason, learn, and make decisions under conditions of uncertainty and resource constraints. By demonstrating aspects of natural intelligence, such as continuous learning and goal-driven behavior, ONA offers a robust framework for developing intelligent systems capable of adapting to the complexities of real-world environments.

\section{Related work}

While we are not aware of any other attempt to integrate functional learning psychology with the Non-Axiomatic Reasoning System (NARS), there are several approaches that aim to implement the “biological basis” of operant conditioning using computational modeling or similar approaches. Importantly, it seems like most of these attempts take a mechanistic approach to operant conditioning, rather than a functional approach (as in this paper). Reinforcement learning, particularly model-free methods like Q-Learning and Deep Q-Networks (DQN), has gained significant attention for its ability to learn optimal policies through interactions with the environment \citep{mnih2015rl}. These methods rely on the Markov property, where the next state depends only on the current state and action, simplifying the learning process but also limiting the system's ability to handle non-Markovian environments.

ONA diverges from RL by adopting a reasoning-based approach grounded in Non-Axiomatic Logic (NAL). Unlike RL, which optimizes a predefined reward function, ONA emphasizes real-time reasoning under uncertainty, adapting to insufficient knowledge and resources \citep{wang2013nalbook}. This allows ONA to handle complex, non-Markovian environments more effectively. While RL methods struggle with sparse rewards and require extensive data to learn, ONA leverages its reasoning capabilities to infer causal relationships and plan actions based on partial knowledge, making it more data-efficient \citep{hammer2022reasoning}.

In summary, while reinforcement learning remains a powerful tool for specific, well-defined tasks, ONA offers a robust alternative for more complex, real-time applications. Its integration of reasoning under uncertainty, goal-driven learning, and adaptability positions it as a significant advancement in the quest for generalizable and resilient AI systems \citep{hammer2022reasoning}.


\section{Machine Psychology}

Machine Psychology is an interdisciplinary framework for advancing AGI research. It aims to integrate principles from operant learning psychology (as described in Section \ref{section:operant}), with the theory and implementation of NARS (as described in Sections \ref{section:nars} and \ref{section:ona}). At the core of the integration is the assumption that adaptation is fundamental to intelligence, both biological and artificial. 

Generally, Machine Psychology can be said to be a \textit{functional} approach (as defined in Section \ref{section:operant} to the problem of building an AGI system. With this, we mean that the Machine Psychology framework enables the possibility to not only study functional relations between changes in the environment and changes in behavior (as in operant psychology), but also to study functional relations between mechanisms and changes in behavior. Hence, both experience of the system, and its mechanisms could in principle be manipulated. 

In the case with studying operant conditioning with NARS, it means that it is indeed possible to both manipulate the system's experience, but also, in principle, to manipulate the mechanisms that are available (or not) during an experimental task. 

This interdisciplinary approach might be likened to Psychobiology, that integrates psychology and biology \citep{dewsbury1991psychobiology}. Psychobiology is an interdisciplinary field that integrates biological and psychological perspectives to study the dynamic processes governing behavior and mental functions in whole, integrated organisms. It emphasizes the interaction between biological systems, such as the nervous and endocrine systems, and psychological phenomena, such as cognition, emotion, and behavior. This approach allows for the dual manipulation of factors related to both experience (as in psychology) and biological processes (as in biology) within a unified framework. By doing so, psychobiology provides a comprehensive understanding of how environmental and experiential factors can influence biological states and how biological conditions can shape psychological experiences, thus bridging the gap between the two domains to offer holistic insights into human and animal behavior \citep{dewsbury1991psychobiology}.

Hence, one way to describe Machine Psychology, is that it is to computer science (and particularly NARS theory), as what Psychobiology is to biology. 

\subsection{An interaction with NARS}

Within a Machine Psychology approach to NARS, it is possible to interact with NARS as one would do with an organism in psychological research in general. This will be illustrated in this section. 

An example interaction can be described as follows, where each line ends with \verb!:|:!, indicating temporal statements. First, a nonsense symbol $A1$ is presented to the left. Then, $A2$ is presented to the right. After that, the event $G$ is established as something desirable by the system (the \verb$!$ indicates that the system desires the event). This triggers the system to execute an operation (for example the operation \verb!^left!). The researcher could then provide the event $G$ as a consequence, leading to a derived contingency statement by the system. The entire interaction can can be described as follows (where \verb!//! represents comments):

\begin{verbatim}
<A1 --> [left]>. :|: // A1 is presented to the left
<A2 --> [right]>. :|: // A2 is presented to the right
G! :|: // G is established as a desired event
^left. :|: // ^left executed by the system
G. :|: // G is provided as a consequence
<(<A1 --> [left]> &/ ^left) =/> G> // Derived by the system
\end{verbatim}

This example aims to provide an example of how the researcher might interact with NARS, as if it was a biological organism. The researchers presents events, and the system responds, and the researcher once again presents an event as a consequence. 

As part of this study, all interactions with ONA was done via its Python interface. Specifically, experimental designs was conducted in the Python-based open source experimental software OpenSesame, that was configured to interact with ONA \citep{mathot2012opensesame}.

\subsection{Learning Psychology with NARS}

Given the above example, it should be clear that interactions between NARS and its environment can be analyzed using the terms provided by functional learning psychology, as provided in Section \ref{section:operant}. The operant learning examples are described at the descriptive level (the level of procedure), and learning effects can be described at the functional level. The analog to the cognitive/mechanistic level is the operations of the NARS system, as for example described in Sedtion \ref{section:ops_ona}. The three-term contingency (as described in Section \ref{section:threeterm}) can be used to describe relations of events, operations and consequences. In the example above, the event \verb!<A1 --> [left]>! functions as a discriminative stimulus, \verb!^left! is a response, and \verb!G! functions as a reinforcer. Importantly though, the \verb$G!$ (that establishes G as as desired event) functions as an establishing operation.

\section{Methods}


\subsection{OpenNARS for Applications}

The study used a version of OpenNARS for Applications (ONA) compiled with the parameter \verb!SEMANTIC_INFERENCE_NAL_LEVEL! set to \verb!0!, which means that only sensorimotor reasoning were to be used. Hence, no declarative inference rules were available during the experiments. 

For all three experiments, ONA was configured at starting time in the following way:

\begin{verbatim}
*babblingops=2
*motorbabbling=0.9
*setopname 1 ^left
*setopname 2 ^right
*volume=100
\end{verbatim}

This indicates that ONA was set to have two operators \verb!^left! and \verb!^right!, and an initial chance of 90\% for motor babbling. 

\subsection{Encoding of experimental setup}

All experimental tasks were presented as temporal Narsese statements, as indicated by the \verb!:|:! markers below. An arbitrary goal event \verb$G! :|:$ was presented at the end to trigger the execution of one of the two procedural operations \verb!^left! and \verb!^right! (through motor babbling or a decision). During training, feedback was given in the form of \verb!G. :|:! (meaning to reinforce a correct choice) or \verb!G. :|: {0.0 0.9}! (to indicate that the system had conducted an incorrect choice). Between each trial, 100 time steps was entered, by feeding \verb!100! to ONA.

\begin{verbatim}
<A1 --> [sample]>. :|:
<B1 --> [left]>. :|:
<B2 --> [right]>. :|:
G! :|:
\end{verbatim}

The first three lines are so-called inheritance statements, with properties on the right-hand side, indicating that the events $A1$, $B1$ and $B2$ are either on the left, right or at the position of a sample. 

\subsection{Experimental designs}

\begin{figure}[h!]
\begin{center}
\includegraphics[scale=0.6]{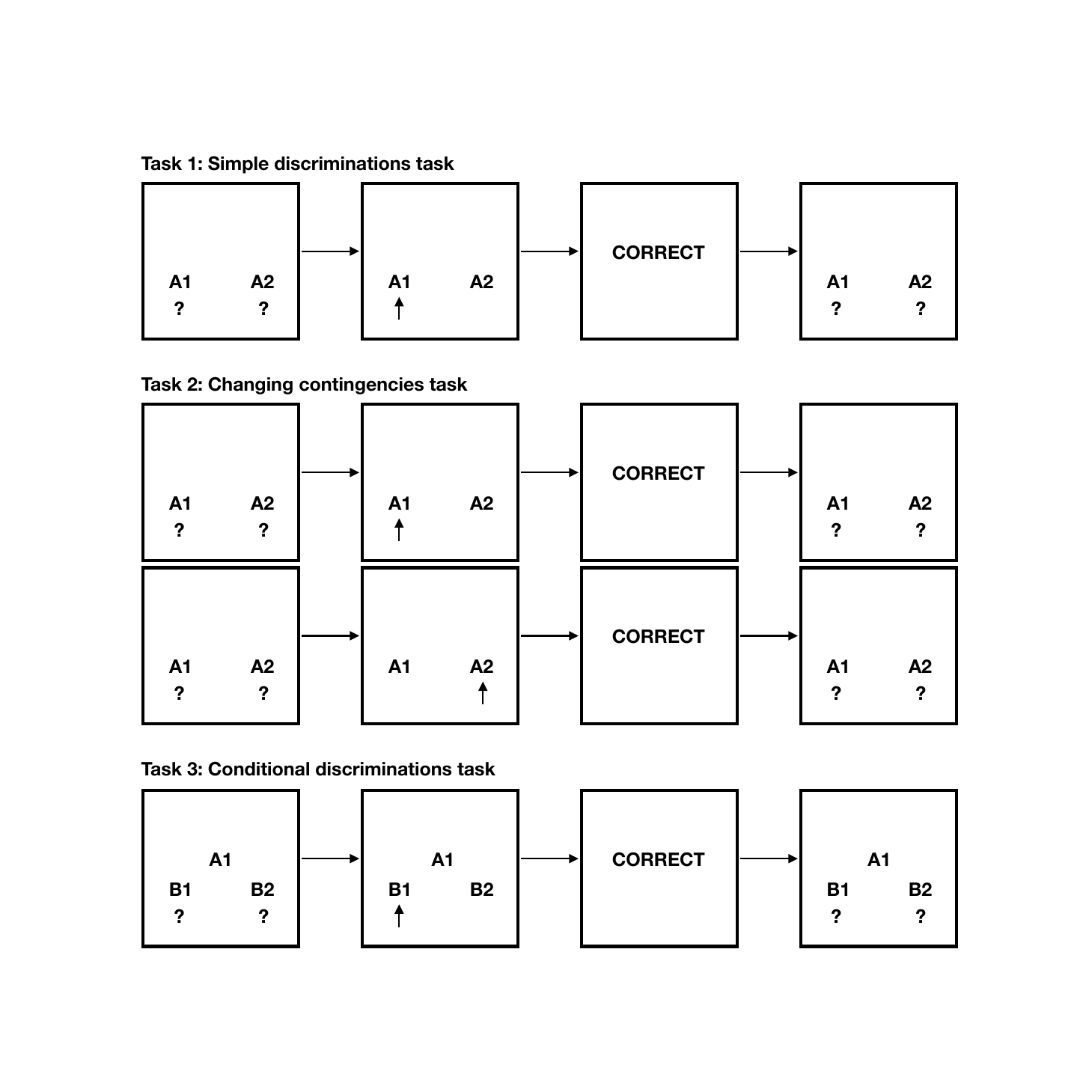}
\end{center}
\caption{
Examples from the three experimental tasks investigated.
} 
\label{fig_tasks}
\end{figure}

In this section, the experimental designs will be detailed of the three tasks: 1) The simple discrimination task, 2) The changing contingencies task, and 3) The conditional discriminations task. The tasks are further illustrated with a few examples in Figure \ref{fig_tasks}. 

The first experiment investigated if NARS could learn in the form of operant conditioning, specifically in the form of simple discriminations. In the experiment, three phases were used: Baseline assessment, Training (with feedback), and Testing (without feedback). In all phases, training and testing were done in blocks of trials. One trial could for example be that $A1$ was to the left, and $A2$ was to the right. A block contained twelve trials, with the two possible trials possible (depending on the location of $A1$ and $A2$), each presented six times in random order.

\begin{enumerate}
	\item \textbf{Baseline: } During the baseline assessment, which was three blocks, no feedback was given. This phase was included to establish a baseline probability of responding correct. It was expected that the system would respond correctly by chance in 50\% of the trials.
	\item \textbf{Training: } Then, the system was trained on a set of three blocks. Feedback was given when the system was correct (for example executing \verb!^left! when $A1$ was to the left), and when not correct. 
	\item \textbf{Testing: } The system was then tested (without feedback) on three blocks, with the contingencies that previously had been trained.
\end{enumerate}

The second experiment investigated if ONA could adapt to changing conditions midway through the task. Five phases were used: Baseline, Training 1 (with feedback), Testing 1 (without feedback), Training 2 (with feedback), and Testing 2 (without feedback). All blocks contained twelve trials.

\begin{enumerate}
	\item \textbf{Baseline: } Two blocks, where no feedback was given. 
	\item \textbf{Training 1: } Four blocks of, where feedback was given. This phase aimed to train the system in executing \verb!^left! when $A1$ was to the left, and \verb!^right! when $A1$ was to the right. 
	\item \textbf{Testing 1: } Then, the system was tested over two blocks (without feedback) on what was trained the previous phase.
	\item \textbf{Training 2: } This phase of four blocks aimed to train in reversed contingencies compared to the first training. That is, the phase aimed to train ONA into executing \verb!^left! when $A2$ was to the left (and hence $A1$ to the right), and \verb!^right! when $A2$ was to the right. 
	\item \textbf{Testing 2: } Over two blocks, the system was tested, without feedback, on the contingencies trained in the previous phase.
\end{enumerate}

Finally, in the third experiment, that investigated if the system could learn conditional discriminations, three phases were used: Baseline, Training, and Testing:

\begin{enumerate}
	\item \textbf{Baseline: } Three blocks of 12 trials, where no feedback was given.
	\item \textbf{Training: } Six blocks, where feedback was given. For example, when $A1$ was the sample, and $B1$ to the left, the system was reinforced for executing \verb!^left!.
	\item \textbf{Testing: } The system was then tested, without feedback, on three blocks of 12 trials, with the contingencies that previously had been trained.
\end{enumerate}



\section{Results}

\subsection{Simple discrimination task}

\begin{figure}[h!]
\begin{center}
\includegraphics[scale=1.0]{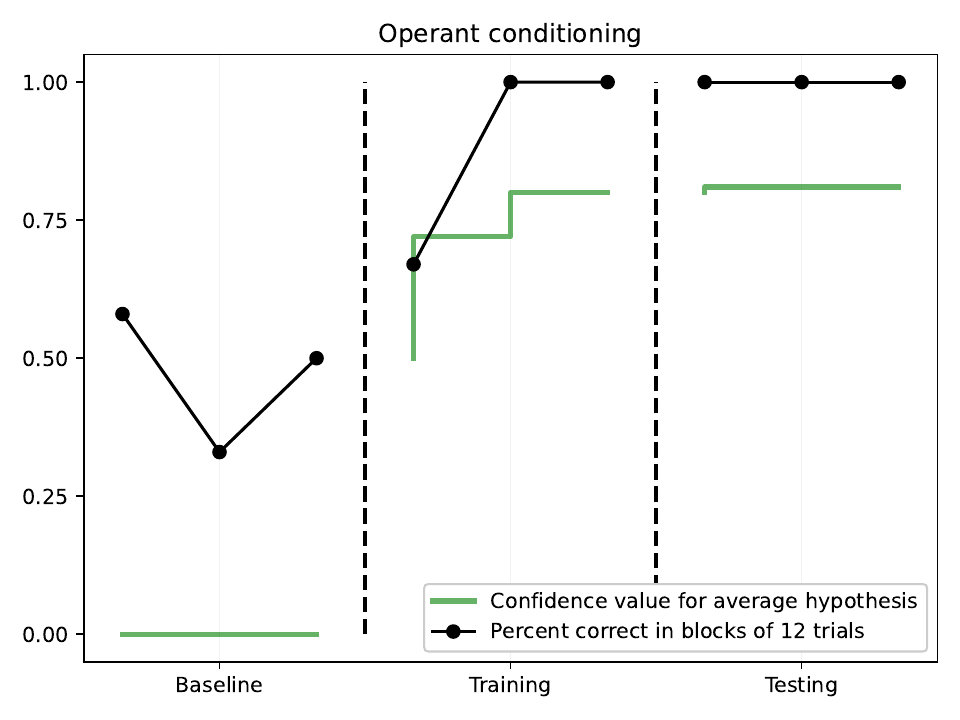}
\end{center}
\caption{
Operant conditioning. Dots illustrate the percent of correct in blocks of 12 trials. The solid line shows the mean NARS confidence value for hypotheses.
} 
\label{fig_results_op1}
\end{figure}

During baseline, the amount of correct trials ranged between 0 and 50\% during the three blocks, indicating that no learning happened. In the training phase, NARS was 100\% correct on all trials already in the second out of three blocks, indicating a rapid learning. Finally, in the testing, where no feedback was provided, NARS performed consistently 100\% correct across all three blocks of trials. The results are illustrated in Figure \ref{fig_results_op1}. 

The average confidence values for the two target hypotheses went from 0.56 to 0.82. These two hypotheses were

\begin{verbatim}
<(<A1 --> [left]> &/ ^left) =/> G>
\end{verbatim}

and 

\begin{verbatim}
<(<A1 --> [right]> &/ ^right) =/> G>
\end{verbatim}

The increase in average confidence value is also illustrated in Figure \ref{fig_results_op1}. 

In summary, the results indicate that ONA indeed can learn in the form of operant conditioning.

\subsubsection{NARS examples from the training phase} \label{section:ex1}

A few example trials from the training session follows. Let’s say that the system was exposed to the following NARS statements:

\begin{verbatim}
<A2 --> [left]>. :|: 
<A1 --> [right]>. :|: 
G! :|:	
\end{verbatim}

If it is early in the training, NARS might use \textit{Motor Babbling} to execute the \verb!^right! operation. Since this is considered correct in the experiment, the feedback \verb!G. :|:! would be given to NARS, followed by 100 time steps. Only from this single interaction, NARS would form a hypothesis using \textit{Temporal Induction}:

\begin{verbatim}
<(<A1 --> [right]> &/ ^right) =/> G>	
\end{verbatim}

When the same situation happens again later during the training phase, ONA will not rely on motor babbling, but instead use its decision making algorithm and \textit{Goal Deduction}, as detailed by \citep{hammer2022reasoning}.

\subsection{Changing contingencies task}

\begin{figure}[h!]
\begin{center}
\includegraphics[scale=1.0]{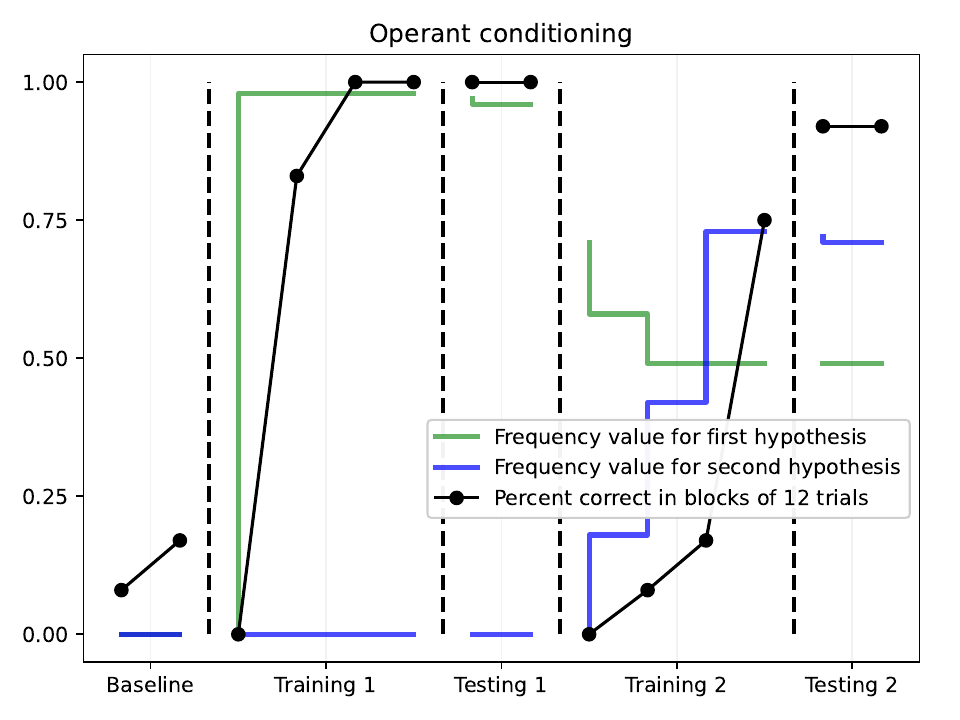}
\end{center}
\caption{
Operant conditioning with changing contingencies. Dots illustrate the percent of correct in blocks of 12 trials. The solid lines show the mean NARS frequency values for the respective hypotheses.
} 
\label{fig_results_op2}
\end{figure}

As expected, no learning happened during the baseline phase, where NARS was less than 25\% correct in both phases. In the first training phase, NARS was 100\% correct after two completed blocks of 12 trials. During testing, the system was 100\% correct without any feedback being present. In the second training phase, where the contingencies were reversed, the system could adapt to the change as indicated by the increase in number of correct responses over time, with 75\% correct in the final block of the phase. Finally, in the second testing phase, the system's performance was 91.7\% correct, indicating that a successful retraining had been conducted. The results are further illustrated in Figure \ref{fig_results_op2}. 

To further illustrate how the NARS system was able to adapt to changing contingencies, the change in average frequency value of the two target hypotheses can be described over time. This is also illustrated in Figure \ref{fig_results_op2}. As seen in Figure \ref{fig_results_op2}, the average frequency value for the first hypothesis was close to 1.0 during the first training and testing, meaning that the system had not received any negative evidence. However, when the contingencies were reversed in the second training phase, the frequency value of the first hypothesis immediately decreased, taking the negative evidence into account. The frequency value of the second hypothesis however, did not rise above zero until the start of the second training, where the hypothesis got positive evidence for the first time. 

These results do all together indicate that a NARS system can adapt in realtime in the form that is necessary when contingencies are reversed midway through a task. 

\subsubsection{Examples from changed contingencies}

The experiment starts out similar as to the example in Section \ref{section:ex1} during the first training phase. However, after the contingencies change, and reinforcement is not provided for executing \verb!^left! and \verb!^right! when $A1$ is to the left and right, respectively, the system is forced to readapt. For example, if the following situation is shown to the system:

\begin{verbatim}
<A1 --> [left]>. :|:
<A2 --> [right]>. :|:
G! :|:
\end{verbatim}

The system will execute \verb!^left! based on its previous learning. However, instead of \verb!G. :|:! as a consequence, \verb!G. :|: {0.0 0.9}! will be provided. An explanation of how \textit{Revision} is used will be provided.

Before the negative feedback, the following hypothesis will have a frequency close to 1.0:

\begin{verbatim}
<(<A1 --> [left]> &/ ^left) =/> G>. {0.98, 0.41}
\end{verbatim}

In the above, \verb!{0.98, 0.41}! means frequency=0.98, and confidence=0.41. 

However, with the negative feedback shown above, the following hypothesis will be derived:

\begin{verbatim}
<(<A1 --> [left]> &/ ^left) =/> G>. {0.00, 0.19}
\end{verbatim}

Together, these two hypothesis with different truth values will be revised as follows:

\begin{verbatim}
<(<A1 --> [left]> &/ ^left) =/> G>. {0.74, 0.48}
\end{verbatim}

When NARS combines the positive and negative evidence, the frequency value goes down from 0.98 to 0.74, and the confidence value goes up from 0.41 to 0.48, as the system has gained even more evidence and is more confident in its conclusions.

With repeated examples smilar to the above, the system will eventually go back to motor babbling, and \verb!^right! will be executed, leading to a reinforcing consequence. That will lead to the following hypothesis being formed:

\begin{verbatim}
<(<A2 --> [right]> &/ ^right) =/> G>.
\end{verbatim}

In summary, the mechanism of \textit{Revision}, in combination to what have been covered previously, enables the system to adapt to changing contingencies.

\subsection{Conditional discriminations task}

\begin{figure}[h!]
\begin{center}
\includegraphics[scale=1.0]{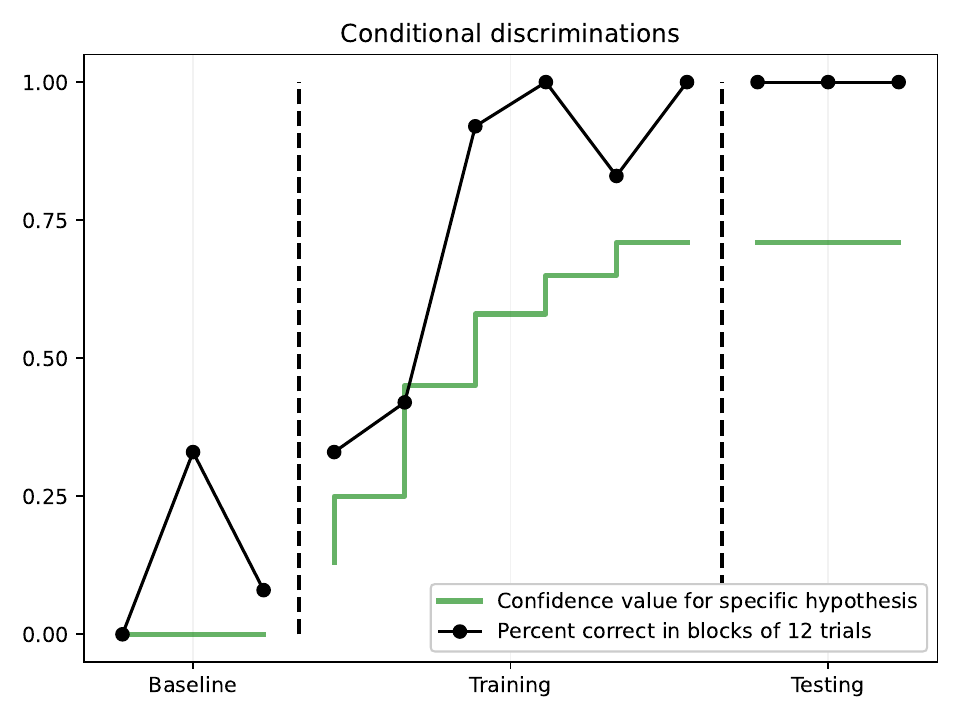}
\end{center}
\caption{
Conditional discriminations. Dots illustrate the percent of correct in blocks of 12 trials. The solid line shows the mean NARS confidence value for hypotheses.
} 
\label{fig_results_cond}
\end{figure}

As with the previous experiments, no learning happened during the three-block baseline. During training, NARS was more than 75\% correct after two completed blocks of 12 trials. In the testing, NARS performed 100\% correct, without feedback, across three blocks of trials. These results are illustrated in Figure \ref{fig_results_cond}. 

The four target hypotheses were the following:

\begin{verbatim}
<((<A1 --> [sample]> &/ <B1 --> [left]>) &/ ^left) =/> G>
<((<A1 --> [sample]> &/ <B1 --> [right]>) &/ ^right) =/> G>
<((<A2 --> [sample]> &/ <B2 --> [left]>) &/ ^left) =/> G>
<((<A2 --> [sample]> &/ <B2 --> [right]>) &/ ^right) =/> G>
\end{verbatim}

The average confidence value for these hypotheses increased from 0.13 to 0.70 during the training phase, as also illustrated in Figure \ref{fig_results_cond}.

\subsubsection{NARS examples from conditional discrimination training}

A few example trials from the training session follows. Let’s say that the system was exposed to the following NARS statements:

\begin{verbatim}
<A1 --> [sample]>. :|:
<B2 --> [left]>. :|:
<B1 --> [right]>. :|:
G! :|:
\end{verbatim}

If it is early in the training, NARS might use motor babbling to execute the \verb!^right! operation. Since this is considered correct in the experiment, the feedback \verb!G. :|:! would be given to NARS, followed by 100 time steps. From this single interaction, NARS would form a hypothesis:

\begin{verbatim}
<((<A1 --> [sample]> &/ <B1 --> [right]>) &/ ^right) =/> G>.
// frequency: 1.00, confidence: 0.15
\end{verbatim}

Importantly, after this single trial, NARS would also form simpler hypothesis such as:

\begin{verbatim}
<(<B1 --> [right]> &/ ^right) =/> G>.
// frequency: 1.00, confidence: 0.21

<(<A1 --> [sample]> &/ ^right) =/> G>.
// frequency: 1.00, confidence: 0.16
\end{verbatim}

This means, that if the same trial was to be presented again (all four possible trials will be presented three times in a block of twelve trials), NARS would respond \verb!^right! again, but the decision being based on the simpler hypothesis, since that hypothesis has the highest confidence value.

Let’s say, that within the same block of 12 trials, the next trial to be presented to NARS was the following:

\begin{verbatim}
<A1 --> [sample]>. :|: 
<B1 --> [left]>. :|: 
<B2 --> [right]>. :|: 
G! :|:
\end{verbatim}

NARS would initially respond \verb!^right!, with the decision being made from the simple hypothesis \verb!<(<A1 --> [sample]> &/ ^right) =/> G>!. 

This would be considered wrong in the experiment, and the feedback \verb!G. :|: {0.0 0.9}! would be given to NARS. This would lead to negative evidence for the simple hypothesis. If the same trial was presented again, NARS would then likely resort to motor babbling that could execute the \verb!^left! operation. Over repeated trials with feedback, the simpler hypotheses would get more negative evidence, and the confidence values of the more complex target hypotheses would increase.

In summary, NARS can learn increasingly complex hypotheses, with repeated examples. 

\subsection{NARS mechanisms}

Given the examples above, we will now provide further clarifications of the results in terms of mechanisms and inference rules that are implemented in ONA. 

In all three tasks, the confidence increase followed from repeated examples which provide evidence to the respective target hypotheses. For this to happen and to derive the truth values, the following mechanisms in NARS were necessary:

\begin{enumerate}
	\item \textbf{Temporal induction: } Given events that $A1$ is to the left, the \verb!^left! operation, and $G$, then derive positive evidence for a relation like \verb!<(<A1 --> [left]> &/ ^left) =/> G>!
	\item \textbf{Goal deducation: } Given for example \verb!<(<A1 --> [left]> &/ ^left) =/> G>! and a precondition that $A1$ is to the left, and the event $G!$, then by deduction derive that the \verb!^left! operation is to be executed.
	\item \textbf{Motor babbling: } The ability to execute operations functions as the means for exploration in the sense that it enables the system to try out new things.
	\item \textbf{Anticipation: } To derive negative evidence to a hypothesis, based on that the antedecent happened but the consequent did not. For example, \verb!<(<A1 --> [sample]> &/ ^right) =/> G>! can receive negative evidence based on anticipation.
	\item \textbf{Revision: } To summarize the positive evidence and the negative evidence for a statement.
\end{enumerate}

\section{Discussion and conclusion}

The results of this study demonstrate the feasibility and effectiveness of integrating principles from operant conditioning with the Non-Axiomatic Reasoning System (NARS) to advance the field of Artificial General Intelligence (AGI). This interdisciplinary framework, referred to as Machine Psychology, offers a novel approach to understanding and developing intelligent systems by emphasizing adaptation, a core aspect of both biological and artificial intelligence.

\subsection{Summary of Findings}

The experiments conducted in this study aimed to evaluate the ability of NARS, specifically the OpenNARS for Applications (ONA) implementation, to perform operant conditioning tasks. The three tasks—simple discrimination, changing contingencies, and conditional discriminations—provided a comprehensive assessment of the system's learning and adaptation capabilities.

In the simple discrimination task, NARS demonstrated rapid learning, achieving 100\% correct responses during the training phase and maintaining this performance in the testing phase without feedback. This indicates that NARS can effectively learn and adapt based on positive reinforcement, a key aspect of operant conditioning.

The changing contingencies task further highlighted the system's adaptability. When the contingencies were reversed midway through the task, NARS was able to adjust its behavior accordingly, showing a significant decrease in errors and an increase in correct responses during the retraining phase. This flexibility is crucial for AGI systems operating in dynamic environments where conditions can change unpredictably.

The conditional discriminations task showcased NARS's ability to handle more complex learning scenarios. Despite the increased difficulty, the system achieved high accuracy, indicating that it can form and utilize more intricate hypotheses based on conditional cues. This capability is essential for developing AGI systems that require sophisticated cognitive skills.

\subsection{Implications for AGI Research}

The success of NARS in these operant conditioning tasks has several important implications for AGI research. First, it validates the use of learning psychology principles, particularly operant conditioning, as a guiding framework for developing intelligent systems. The results suggest that mechanisms enabling operant conditioning are integral to the development of adaptive behaviors and cognition in AGI systems.

Second, the experiments carried out as part of this study, can be said to constitute key milestones of AGI research, as has been suggested by us elsewhere \citep{johansson2020scientific}. Operant psychology research provides examples of increasingly complex tasks, that can be used to test the abilities of an AGI system. The use of functional learning psychology principles to guide AGI research also enable metrics to be used to evaluate AGI systems, as demonstrated in this paper.

Third, the study highlights the potential of NARS as a robust model for AGI. Unlike traditional AI systems that rely on predefined algorithms and large datasets, NARS operates effectively under conditions of insufficient knowledge and resources. This adaptability makes it well-suited for real-world applications where information is often incomplete and environments are constantly changing.

Fourth, the integration of sensorimotor reasoning with operant conditioning principles in ONA provides a scalable and flexible framework for AGI development. By demonstrating aspects of natural intelligence, such as continuous learning and goal-driven behavior, ONA offers a practical approach to building intelligent systems that can interact with and learn from their environments in real-time.

\subsection{Future Directions}

The findings of this study open several avenues for future research. One potential direction is to explore the integration of additional cognitive and behavioral principles from functional learning psychology into NARS. Future research can be guided by operant theories of cognition, such as Relational Frame Theory, as suggested by \cite{johansson2019aarr}.

Another important direction is to apply the Machine Psychology framework to more complex and diverse tasks beyond the idealized examples provided in this paper. By testing NARS in various real-world scenarios, such as autonomous robotics, natural language processing, and human-computer interaction, researchers can evaluate the system's generalizability and robustness across different domains.

Additionally, further refinement of the sensorimotor inference and declarative inference components in ONA could lead to improvements in the system's performance. Enhancing the efficiency of resource management, memory structures, and event-driven control processes will be critical for scaling up the system to handle more sophisticated tasks and larger datasets.

\subsection{Conclusion}

In conclusion, this study demonstrates that integrating operant conditioning principles with NARS offers a promising pathway for advancing AGI research. The Machine Psychology framework provides a coherent and experimentally grounded approach to studying and developing intelligent systems. By emphasizing adaptation and learning from environmental interactions, this interdisciplinary approach has the potential to significantly advance the field of AGI and bring us closer to achieving human-level artificial intelligence.

\section*{Conflict of Interest Statement}

The authors declare that the research was conducted in the absence of any commercial or financial relationships that could be construed as a potential conflict of interest.

\section*{Author Contributions}

RJ solely conceived and designed the study, performed the experiments, analyzed the data, and wrote the manuscript.


\section*{Funding}
This research received no specific grant from any funding agency in the public, commercial, or not-for-profit sectors. However, the author is funded by Digital Futures, Stockholm, Sweden and would like to acknowledge them for financial support.


\section*{Acknowledgments}
The author would like to thank Patrick Hammer and Tony Lofthouse for many valuable discussions regarding the work presented in this paper.

\bibliographystyle{apalike}
\bibliography{references}  

\section*{Figure captions}



\end{document}